\documentclass[letterpaper]{article}
\usepackage{aaai}
\usepackage{url}

\usepackage{amsmath, amssymb}
\usepackage{tikz}
\usepackage{pgfplots}
\usepackage{booktabs}
\usepackage{gensymb}
\pgfplotsset{compat=1.14}
\usepgfplotslibrary{fillbetween}

\newcommand{\cloud}{c}
\newcommand{\astep}{t}
\newcommand{\amesh}{m}
\newcommand{\allmeshes}{\mathcal{M}}
\newcommand{\nmeshes}{K}

\newcommand{\maxsteps}{T}

\newcommand{\astate}{s}
\newcommand{\aaction}{a}

\newcommand{\pstate}{\mathcal{S}}
\newcommand{\paction}{\mathcal{A}}
\newcommand{\pproba}{P}
\newcommand{\preward}{R}
\newcommand{\pobs}{O}

\copyrighttext{\copyright\space \copyright@year{} All rights reserved.}

\begin{document}
%
\title{Schedule Earth Observation satellites with Deep Reinforcement Learning}
\author{Adrien Hadj-Salah\textsuperscript{1,2},\quad
R\'{e}mi Verdier\textsuperscript{1},\quad
Cl\'{e}ment Caron\textsuperscript{1,2},\quad
Mathieu Picard\textsuperscript{1,2},\quad
Mika\"{e}l Capelle\textsuperscript{1}\\
\textsuperscript{1}IRT Saint-Exup\'{e}ry \quad
\textsuperscript{2}Airbus Defence \& Space\\
\{adrien.hadj-salah, remi.verdier, clement.caron, mathieu.picard, mikael.capelle\}@irt-saintexupery.com
}
\maketitle
\begin{abstract}
\begin{quote}
Optical Earth observation satellites acquire images world-wide, covering up to several million square kilometers every day. The complexity of scheduling acquisitions for such systems increases exponentially when considering the interoperability of several satellite constellations together with the uncertainties from weather forecasts. 
In order to deliver valid images to customers as fast as possible, it is 
crucial to acquire cloud-free images. 
Depending on weather forecasts, up to 50\% of images acquired by operational 
satellites can be trashed due to excessive cloud covers,
showing there is room for improvement.
We propose an acquisition scheduling approach based on Deep Reinforcement Learning and experiment on a 
simplified environment. We find that it challenges classical methods relying on human-expert heuristic.
\end{quote}
\end{abstract}

\section{Introduction}

Earth Observation (EO) systems acquire cloud-free images and deliver them to 
customers worldwide on a daily basis.
Requests come in a variety of size and constraints, from the 
urgent monitoring of small areas to large area coverage. 
In this work we are particularly interested in the latter case, with 
requests covering whole countries or even continents.
Depending on weather conditions, such requests may take several months to complete, 
even with multiple satellites.

In order to shorten the time required to fulfill requests, the mission orchestrator
shall schedule acquisitions with both a short and a long-term strategy. 
Determining a strategy robust to an uncertain environment is a complex task, this is why current solutions mainly consist of heuristics configured by human-experts.
This paper demonstrates that Reinforcement Learning (RL) might be well-suited for
such a challenge. 
RL has proven to be of great value since these algorithms have
mastered several games such as Pong on Atari 2600~\cite{mnih2013playing}, Go with AlphaGo~\cite{silver2017mastering} and more recently Starcraft~\cite{arulkumaran2019alphastar}. 

\section{Scheduling acquisitions for Earth observation systems}

\subsection{Single satellite acquisition scheduling}

EO satellites carry optical instruments which are able to take acquisitions with a specific width, called swath, and a maximum length depending on the satellite models.
The capacity of the satellites to take multiple images along their orbit track is related to their agility~\cite{lemaitre2002selecting}.

Due to limited swath and acquisition length, a large area must be split into tiles 
called meshes. 
For instance, considering the Pleiades satellites, covering France requires thousands of meshes.
A satellite overflying an area is able to acquire a sub-part of those meshes
due to its limited agility. 
With sun-synchronous orbit, revisit of a ground point takes days  
which explains the importance of mesh selection~\cite{gleyzes2012pleiades}.
  
The satellite schedule is computed on ground by the Mission Planning Facility (MPF), where an optimization algorithm selects the top-ranked acquisitions and ensures the kinematic feasibility of the attitude maneuvers.

\subsection{Interoperable EO systems scheduling for large-area coverage}

The trend of EO systems is toward large constellations of heterogeneous satellites. For instance, Airbus Intelligence, operating the well-known SPOT and Pleaides satellites, will soon manage a new system of 4 satellites (Pleaides NEO). Dealing with multiple EO systems needs both human expertise and algorithms to dispatch requests over the satellites and to deliver end customers on time. 

We approach the constellation scheduling by having an orchestrator responsible for request ranking towards each MPF. The orchestrator analyzes a large-scope of data (e.g., forecasts, access opportunities) to optimize the global schedule, while each MPF has a narrowed and short term vision of their single (or dual) satellite scheduling. Additionally, we focus in this paper on requests consisting in a large area (countries, continents). Such requests usually contain several hundreds of meshes to acquire over long periods (up to several months).

The two main contributors to the overall uncertainty on the time to completion are: firstly the weather conditions at the time of acquisition, which can only be forecasted, and secondly the presence of other requests within the systems, arriving at an unknown rate.

This explains our focus on RL algorithms which have the capacity to learn new strategies, robust to uncertainties, while challenging traditional approaches.

\section{Reinforcement Learning approach}

In Reinforcement Learning, an agent learns how to behave through
trial-and-error interactions with a dynamic environment.
The actions the agent takes are decided by a \textit{policy}, which
can be seen as a function mapping information from the environment
to actions.
The goal of reinforcement learning algorithms is to find an optimal
policy, i.e., a policy that maximizes the reward of the agent over
the long-term.

Recently, deep neural networks have proven to be efficient
for finding policies.
Several deep-RL algorithms are actively studied to solve complex sequential decision-making problems. 
Among the best-known methods, one can cite value-based algorithms such as
DQN, Rainbow~\cite{hessel2018rainbow}, policy-based algorithms such as 
REINFORCE~\cite{sutton2000policy} or actor-critic methods such as A2C~\cite{mnih2016asynchronous} 
or PPO~\cite{schulman2017proximal}.

\subsection{Problem simplification}\label{ssect:problem_simplification}

In order to evaluate the benefits of Reinforcement Learning,
we propose a simplified environment.

We consider that all satellites have the same swath, thus the  tessellation
(i.e., the meshes) of the area is the same for all satellites. 
We also assume that each satellite can acquire at most one mesh per pass over the considered area. 
A satellite pass occurs when it overflies the large-area request on a given orbit.
The planned mesh is validated or rejected depending on actual
cloud cover observations at the time of acquisition. We do not consider uncertainties related to the load of our system,
i.e., satellites are always fully available.

The area of interest (AOI) is enclosed in a rectangular box –-- considering a Mercator projection –-- containing $N_{lat}\times{}N_{lon}$ meshes.
Since some meshes of this grid mesh may not belong to the AOI, we define $\allmeshes = \{\amesh_k~:~1 \leq k \leq \nmeshes\}$, the set of meshes to acquire.

For each pass $\astep\in\mathbb{N}$, we denote by $\allmeshes_\astep\subseteq\allmeshes$
the subset of meshes in the AOI that can be acquired by the corresponding satellite knowing its orbit and agility.

We denote by $\cloud^a_\astep(\amesh)$ and $\cloud^f_\astep(\amesh)$ the
actual and forecast cloud cover above mesh $\amesh$ during pass $\astep$.

\subsection{Problem formulation}\label{ssect:problem_formulation}

The given problem can be formalized as a \textbf{Markov Decision Process} (MDP) 
which is an intuitive and fundamental formulation for RL~\cite{bensana1999dealing}.

An agent interacts with the environment by taking actions from a legal set of actions. The agent purpose is to acquire $\allmeshes$ as quickly as possible. For each step $\astep$, only one mesh can be selected. The chosen mesh is then validated or rejected depending on weather conditions. 

The state space $\pstate{}$, the discrete action space $\paction{} \subset \mathbb{N}$, the stochastic discrete-time transition function $\pproba{}$ and the reward function $\preward{}$ define the underlying MDP: $M = <\pstate{}, \paction{}, \pproba{}, \preward{}>$.

The horizon is considered finite. Therefore, there is a finite number of discrete time steps $\astep$ during an episode. Each episode comprises a maximum of $\maxsteps \in \mathbb{N}^*$ steps. The state space $\pstate$ is defined as:

\begin{equation*}
    \pstate = \pstate_{status} \times \pstate_{time} \times \pstate_{passes}
\end{equation*}
where $\pstate_{status} = \{0,~1\}^{N_{lat}\times N_{lon}}$ characterizes the status of each mesh (i.e., already validated or to acquire),
$\pstate_{time} \subset \mathbb{R}$ encodes the date of the current satellite pass $\astep$ and
$\pstate_{passes} \subset \mathbb{R}^d$  describes all pass dates,  accessible meshes $\allmeshes_t$ and related weather forecasts.

The goal is to find a policy $\pi: \pstate \rightarrow \paction$ that maximizes the expected discounted reward over the finite horizon:
\begin{equation}\label{eq:discounted_reward}
    \sum_{\astep=0}^{\maxsteps} \gamma^\astep 
    \preward(\astate_{\astep}, \pi(\astate_{\astep}), \astate_{\astep + 1}) 
\end{equation}
where $0 \leq \gamma < 1$ is the discount factor and $\astate \in \pstate$.

\subsection{Action space}

At each discrete step $\astep$, the learning agent takes an action. A step corresponds to a satellite pass over the AOI and the action is to pick up a single mesh to acquire during this pass.
\begin{equation*}
\paction = \{0,~1, \ldots ,~K\}
\end{equation*}
We denote $\aaction_k$ the action selecting the mesh $\amesh_k$.
Note that $|\paction| = K +1$ because there is one more ``do nothing'' action available for the agent.

\subsection{Observation space}

At a given step, the agent perceives only useful and available information about the environment. The problem is generalized to a Partially Observable Markov Decision Problem (POMDP).

The observation space $\pobs$ provides information about the mesh status and their validation probability for the following $N_{pass}$ passes, including the current pass for which the agent shall select a mesh.
The validation probability of a mesh depends on weather forecast accuracy, as detailed in Section~\ref{ssect:transition-function}.
Thus, an observation is a tensor with a shape $(N_{lat},~N_{lon},~N_{pass}+1)$.

The observation can be seen as a stack of $N_{lat} \times N_{lon}$ matrices. Each frame (i.e., 2D matrix) encodes information for all tiles of the grid mesh.
This representation preserves spatial information and enables the use of Convolutional Neural Networks.

The validation frame encodes the status of each mesh:  validated ($0$) or to be validated ($1$). We denote the validation frame space $\pobs_{status} = \pstate_{status}$.

The validation probability frames belong to the space
$\pobs_p~ = ~[0,~1]^{N_{lat} \times N_{lon} \times N_{pass}}$.
They encode the probability $p_t$ to acquire and validate 
each mesh for each pass in chronological order from time step $\astep$.
For a given mesh $\amesh$ and a given pass $n \in \{1, \ldots, N_{pass}\}$ at the step $\astep$:

\begin{equation*}
p_t(m, n) = \left\{ 
\begin{array}{ll}
     0 \quad \text{if }\amesh\notin{}\allmeshes_{t} \\
     \mathbb{P}(\cloud^a_{t_n}(m) \leq \cloud_{max}~|~\cloud^{f}_{t_n}(m)) \quad \mbox{otherwise}
\end{array}
\right.
\end{equation*}
with $\cloud_{max}$ the total cloud cover validation threshold.
$t_n = t + n - 1$ is the time related to the pass $n$ knowing that the current time step is $t$.

We can now define $\pobs = \pobs_p \times \pobs_{status}$

\subsection{Reward}

A reward is given to the agent at each step. The value of the reward depends on the status of the chosen mesh before and after this step. $\preward: \pstate \times \paction \times \pstate \rightarrow \mathbb{R}$ gives rewards for particular transitions between states.

\begin{equation*}
\preward(\astate_{\astep}, \aaction_k, \astate_{\astep + 1}) = \left\{ 
\begin{array}{ll}
     1 & \mbox{if $\amesh_k$ is newly validated} \\
     0 & \mbox{otherwise}
\end{array}
\right.
\end{equation*}
This dense reward encourages the agent to reduce the completion time with a discount factor $\gamma<1$ (\ref{eq:discounted_reward}).

\subsection{Transition function}\label{ssect:transition-function}

$\pproba: \pstate \times \paction \times \pstate \rightarrow [0,~1]$ denotes the transition function.
\begin{equation*}
 \pproba(\astate_{\astep}, \aaction, \astate_{\astep+1}) = \mathbb{P}(\astate_{\astep+1} \vert  \astate_{\astep}, \aaction)   
\end{equation*}

For each transition, the current state is updated. $\astate^{time} \in \pstate_{time}$  takes the value of the next pass date in the chronological order. $\astate^{passes} \in \pstate_{passes}$ remains the same during the whole episode. $\astate^{status} \in \pstate_{status}$ is updated if the selected mesh is validated:
\begin{equation*}
\mathbb{P}(\astate^{status}_{t+1}(\amesh_k) = 0 ~\vert~ \astate^{status}_{\astep}(\amesh_k) = 1, \aaction_{k}) = p_t(\amesh_k, n)
\end{equation*}
where $\astate^{status}_{\astep}(\amesh_k)$ is the status of $\amesh_k$ at $\astep$.\\
This probability is computed considering the following weather model:
\begin{equation*}
\cloud^a_{t}(m) = \cloud^f_{t}(m) + \chi(m)
\end{equation*}
\begin{equation*}
\begin{array}{ll}
     \text{with} & \chi(m)\sim\mathcal{N}(\cloud^f_{t}(m),\sigma(\cloud^f_{t}(m))^{2}) \\
     \text{and} & 
     \sigma(x) = u \times x + v
\end{array}
\end{equation*}
$\sigma$ is a linear function computing a representative deviation between forecast and observed data.

\section{Experiments}

Based on the hypotheses from Section~\ref{ssect:problem_simplification}, we implement a simulator using the OpenAI Gym framework. 

\subsection{Scenario}

To evaluate the agents, we choose mainland France as our area of interest. 
It is an interesting case to study because one mesh selection can have an important impact on the mission length due to the territory climate variability. 

We consider 4 satellites with a common 60 km swath, implying $\nmeshes=122$ meshes for our
tesselation.
Each scenario begins at a random date.

We use the ERA-Interim dataset~\cite{dataset:erai} which provides total cloud cover observations on a $0.5\degree\times{}0.5\degree$ grid to compute the observation space.
In the weather model, $u$ is fixed to 0.1 and $v$ to 0.2. 
$\cloud_{max}$ is set to 20\% for all scenarios.

\subsection{Reference algorithms}

In order to benchmark the performances of our agent, we define a random agent and a heuristic that selects one mesh among $\allmeshes_\astep$ for each time step $\astep$:

\begin{itemize}
\item \textbf{Random} that selects the mesh randomly among accessible meshes at each pass.
\item \textbf{Heuristic} that selects the mesh with the highest trade-off score 
between short-term and long-term probabilities $p_{t}$:
    \begin{equation*}
        p_t(m, 1) + \alpha\left(1-\frac{1}{N_{pass}-1}\sum_{n=2}^{N_{pass}} \beta^n p_t(m,n)\right)
    \end{equation*}
    where $\alpha$ is the weight on future passes and $\beta$ the 
    discount factor that favors near future passes. The best performances are reached with ($\alpha=1$, $\beta=0.99$) for $N_{pass}=20$. 
\end{itemize}

\subsection{Train and test methodology}

To avoid overfitting, we use a train and test split methodology on 
the weather data.
Training is done using data from the years 2013 and 2014, 
while testing is done with data from 2015.

We concentrate our experiments on the A2C algorithm which gives the best results. We train A2C agents using two observation spaces:
one with a short-term vision ($N_{pass}~=~1$) and one with a long-term vision ($N_{pass}~=~20$). 
Those A2C agents are respectively named A2C-1 and A2C-20. 
We use the A2C implementation from the OpenAI baselines framework~\cite{baselines}
and train agents during $3\times{}10^7$  steps using 16 parallel environments 
($\sim{}30$ hours using a K80 GPU and 8 vCPUs).
Other hyper-parameters are set to default values.

We use a neural network architecture made of a convolution block followed
by a dense block with two heads: one to estimate the state value and one
to estimate the policy distribution.
The convolution block contains three convolutional layers with decreasing kernel sizes ($7\times7$, $3\times3$, $1\times1$), $128$ filters per layer and ReLU activation functions.
The value and policy heads are only made of a dense layer with respectively one 
unit and $\nmeshes+1$ units.

\begin{figure}
  \centering
  \includegraphics[width=\linewidth]{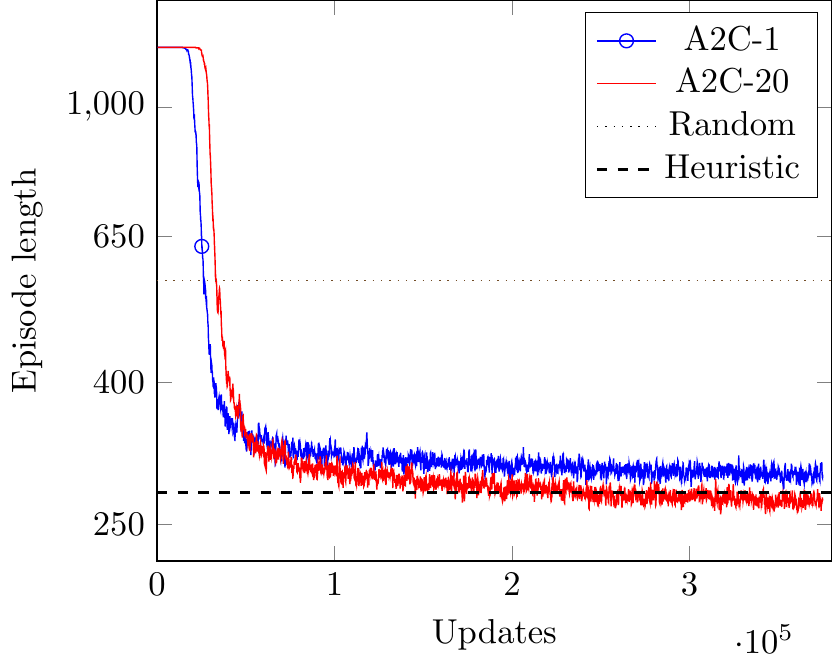}
  \caption{Episode mean length for the last 100 episodes of each 
  training phase.}\label{figure:training}
\end{figure}

Figure~\ref{figure:training} shows the mean length of the last 100 episodes as a function of the number of network weight updates for A2C-1 and A2C-20. 
In our environment, the length of an episode directly relates to the completion time of the area.
We set a maximum number of $10\times\nmeshes$ time steps before resetting the environment to avoid too long episodes when the policy does not perform well. The performances of the trained agents converge close to the heuristic one. Best results are achieved with A2C-20.

During testing phase, we select days from 2015 as starting dates. 
For each date we assess the performances of the models and the reference algorithms. 
We repeat the operation using 3 different weather seeds ($3 \times 365$ runs in total). 
Table~\ref{table:result1} presents statistics on the episode length for the different agents. 
We find that for both agents the transfer on the new weather data went well. 
\begin{table}
\centering
\begin{tabular}{rccr}
\toprule
Agent & Mean & Median & Std\\
\midrule\\
Random & 568.8 & 572 & 110.5\\
Heuristic & 292.7 & 298 & 56.0\\
A2C-1 & 299.3 & 304 & 58.2\\
A2C-20 & 278.5 & 281 & 55.8\\
\bottomrule
\end{tabular}
\caption{Mean, median, standard deviation of the episode lengths
for the different agents.}\label{table:result1}
\end{table}

A2C-20 still provides the best results winning the heuristic in almost 80\% of the cases. It confirms the intuition that a long term strategy is necessary to optimize time-to-completion. 

\section{Conclusion}

This paper demonstrates how Reinforcement Learning can be used in Earth Observation satellites scheduling in order to reduce the time-to-completion of large-area requests. The computed network has been trained to rank the requests and dispatch them to the satellites. In a series of simulation-based experiments, the proposed method challenges the state-of-the-art heuristics.

In future research, we aim to improve the simulation representativeness in order to pave the way for a potential industrial transfer.

\bibliographystyle{aaai}
\bibliography{references}{}

\end{document}